\documentclass[journal,twoside,web]{ieeecolor2}
\usepackage{generic}
\usepackage{cite}
\usepackage{amsmath,amssymb,amsfonts}
\usepackage{algorithmic}
\usepackage{graphicx}
\usepackage{textcomp}
\def\BibTeX{{\rm B\kern-.05em{\sc i\kern-.025em b}\kern-.08em
    T\kern-.1667em\lower.7ex\hbox{E}\kern-.125emX}}
\begin{document}
\title{Unsupervised Domain Adaptation for Calcification Classification in Mammography Across Multi-Site Datasets}
\author{Xuan Liu, Derek L. Nguyen, Emily C. Barre, Jennifer Thomas, Thomas Lynch, Jeffrey R. Marks, E. Shelley Hwang, Marc D. Ryser, Joseph Y. Lo, Lars J. Grimm 
\thanks{This study was supported in part by NIH/NCI R01 CA271237. The mammography images and data used in this research were derived from the OPTIMAM imaging database (OMI-DB) (https://medphys.royalsurrey.nhs.uk/omidb). We would like to acknowledge the OPTIMAM project team and staff at the Royal Surrey NHS Foundation Trust who developed the OPTIMAM database, Cancer Research UK which funded the creation and maintenance of OPTIMAM database and Cancer Research Horizons which facilitates access to the OPTIMAM data. (Corresponding author: Lars J. Grimm)}
\thanks{Xuan Liu is with the Department of Electrical and Computer Engineering and the Department of Radiology, Duke University, Durham, NC, USA (e-mail: xuan.liu115@duke.edu).}
\thanks{Derek L. Nguyen, Joseph Y. Lo and Lars J. Grimm are with the Department of Radiology, Duke University, Durham, NC, USA (e-mail: derek.nguyen@duke.edu; joseph.lo@duke.edu; lars.grimm@duke.edu).}
\thanks{Emily C. Barre is with the Duke University School of Medicine, Durham, NC, USA (e-mail: emily.barre@duke.edu).}
\thanks{Jennifer Thomas is with the Department of Population Health Sciences, Duke University, Durham, NC, USA (e-mail: jennifer.grant@duke.edu).}
\thanks{Thomas Lynch, Jeffrey R. Marks and E. Shelley Hwang are with the Department of Surgery, Duke University, Durham, NC, USA (e-mail: thomas.lynch2@duke.edu; jeffrey.marks@duke.edu; shelley.hwang@duke.edu).}
\thanks{Marc D. Ryser is with the Department of Population Health Sciences, Duke University, Durham, NC, USA and the Faculty of Medicine, University of Geneva, Geneva, Switzerland (e-mail: marc.ryser@duke.edu).}
}

\maketitle

\begin{abstract}
Deep learning-based computer-aided diagnosis (CAD) systems have shown strong performance in breast cancer diagnosis, particularly for classification tasks in mammography. However, domain shifts across multi-site datasets remain a challenge, especially when models are applied to unseen domains. In this work, we proposed a calcification classification framework to improve malignant versus benign breast disease classification across multi-site mammography datasets. The framework consisted of two components: (1) an unsupervised domain adaptation module based on style transfer models (AdaIN and CycleGAN) to generate vendor-specific and technique-specific training samples without additional annotations, and (2) a supervised classification module using Swin Transformer V2 as the backbone. We evaluated the proposed method on three datasets: cross-validation on OPTIMAM (National Health Service, United Kingdom; n=2994), followed by external validation on EMBED (Emory University; n=125), and Duke Calcification Dataset v1 (n=788). These datasets cover multiple vendors and include both full-field digital mammography and synthetic 2D images derived from digital breast tomosynthesis. The proposed framework improved cross-site performance for both EMBED (AUC 0.68 to 0.72) and the Duke Calcification Dataset (AUC 0.68 to 0.73). These findings indicate that domain adaptation can reduce domain shifts and improve the generalization for calcification classification across multi-site datasets. 
\end{abstract}

\begin{IEEEkeywords}
Classification, Domain Adaptation, Style Transfer, Mammography, Computer-aided Diagnosis
\end{IEEEkeywords}

\section{Introduction}
\label{sec:introduction}
\IEEEPARstart{B}{reast} cancer is one of the leading causes of cancer-related deaths among women in the United States, highlighting the urgent need for early detection and improved diagnostic methods to reduce mortality \cite{b1}. Mammography is the primary breast cancer screening modality and has proven effective for early diagnosis and improved prognosis \cite{b2}. Breast cancer can present in many different forms on mammography, including both soft tissue lesions (e.g., mass, asymmetry) and calcifications. Calcifications can be very small and highly variable in size, shape, and distribution, which presents unique challenges compared to other lesion types \cite{b1}. Calcifications are common mammographic findings and typically benign, although some calcifications can represent pre-cancerous disease or an early form of breast cancer, particularly ductal carcinoma in situ (DCIS), so accurately predicting their underlying pathology is critical \cite{b14}.

In recent years, computer-aided diagnosis (CAD) systems, especially those based on artificial intelligence and deep learning, have achieved remarkable advances in breast cancer detection \cite{b3,b4,b5}, segmentation \cite{b6,b7} and classification \cite{b8,b9,b10}. However, most existing CAD studies are limited by evaluating only small, private, single-site datasets. These limitations persist because of a scarcity of large, publicly available datasets and the difficulty of obtaining curated annotations for each image. Furthermore, calcifications are less common than soft tissue lesions and are therefore underrepresented in available datasets and have received less emphasis in CAD model development.

To compensate for these limitations, datasets from different sources can be collated for analysis. However, the inclusion of data from multiple sites introduces domain shifts arising from several sources: (1) scanner hardware, (2) annotation, (3) imaging technique, (4) acquisition protocol, and (5) patient cohort differences \cite{b11,b15}. These cross-dataset domain shifts can substantially degrade CAD performance when models are applied to data from previously unencountered vendors and institutions \cite{b11,b12,b13}. This issue is particularly noticeable in deep learning-based models that rely on large training data but often show limited generalization to new domains, a common scenario in medical imaging. Expanding existing or adding new datasets is costly, time-consuming and requires substantial labeling effort, so effective domain adaptation strategies are needed to reduce performance gaps.

In this work, we focused on addressing domain shifts using data from three sources. Specifically, we aimed to mitigate discrepancies between common mammography vendors (Hologic and GE) and two routine image acquisition techniques (conventional 2D full field digital mammography (FFDM) and synthetic 2D images derived from 3D digital breast tomosynthesis) for the analysis of calcifications. We adopted two style transfer-based unsupervised domain adaptation methods, AdaIN \cite{b16} and CycleGAN \cite{b17}, to perform image-to-image translation between source and target domains. These methods generated targe-style images that could be used as additional training data without requiring extra annotations. We hypothesized that incorporating such data could effectively reduce domain gaps and improve downstream calcification classification performance in multi-site evaluations.

\begin{figure}[!t]
\centerline{\includegraphics[width=\columnwidth]{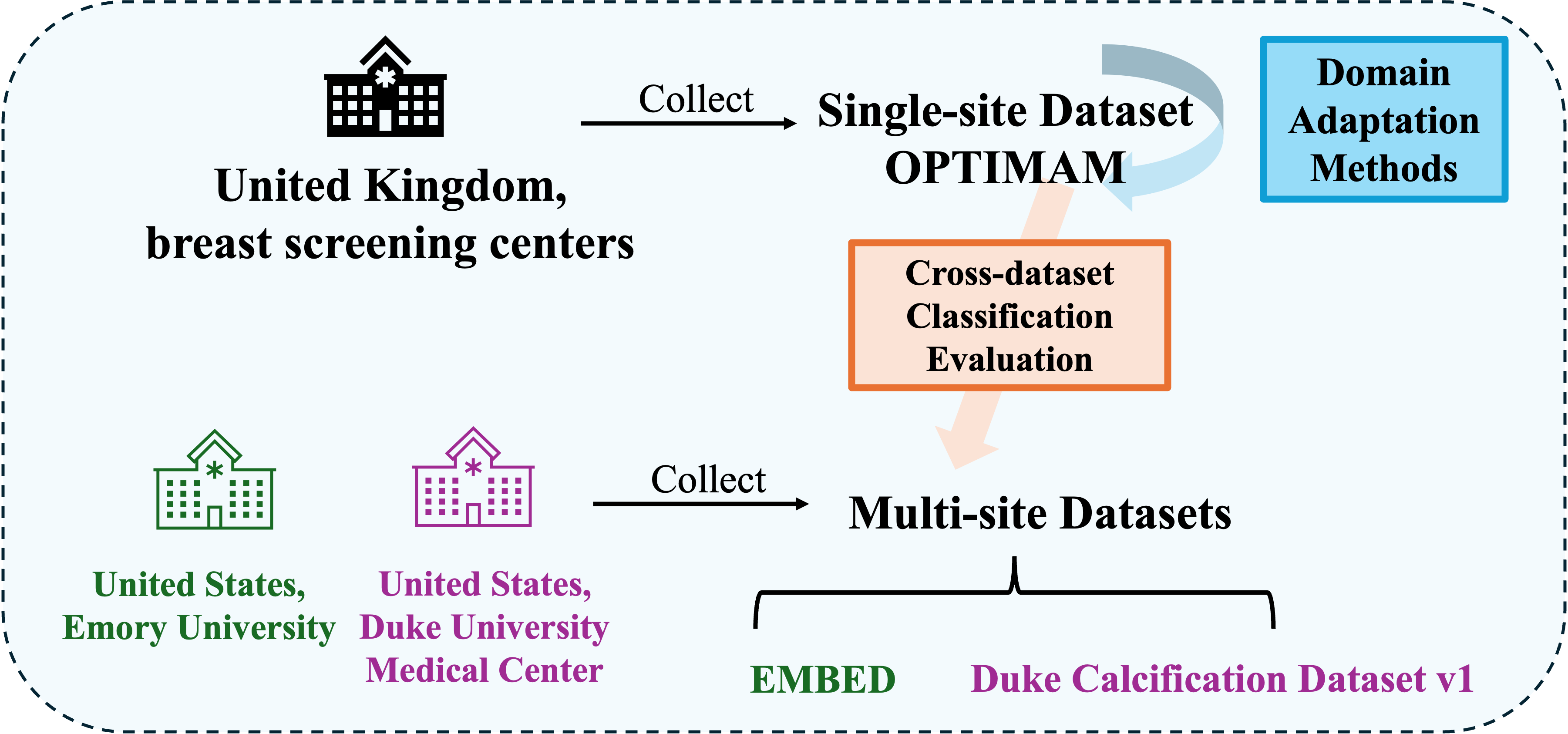}}
\caption{Illustration of single-site and multi-site dataset settings and domain adaptation across institutions.}
\label{fig1}
\end{figure}

The main contributions of this work are summarized as follows:
\begin{enumerate}
    \item We first compared calcification classification performance (benign vs. malignant) across multiple representative backbone networks on the single-site OPTIMAM dataset \cite{b18}, and selected Swin Transformer V2 \cite{b55} as the optimal baseline backbone for subsequent experiments. 
    \item Then, to improve generalization across multi-site datasets, we proposed a multi-stage classification framework consisting of an unsupervised domain adaptation module that generates vendor- and technique-specific patches without requiring additional annotations, followed by a cascade classification module that outperforms the baseline across multi-site datasets (Figure \ref{fig1}).
    \item Finally, we evaluated the proposed framework on two public (OPTIMAM \cite{b18}, EMBED \cite{b19}) and one private dataset (Duke Calcification Dataset v1: 2026) with curated lesion-level annotations, including 2994, 125 and 788 cases, respectively. For the domain adaptation module, we used over 1000 unlabeled mass cases from two different vendors, including both FFDM and synthetic images.
\end{enumerate}

To the best of our knowledge, this is the first work to systematically analyze and address the domain shifts across multi-site datasets for the classification of benign versus malignant calcifications. 

\section{Related Work}
\subsection{Deep Learning-Based Breast Cancer Classification}
Breast cancer classification is a fundamental task in mammogram analysis. Deep learning approaches, including both convolutional neural network (CNN)-based models \cite{b20, b21,b22} and transformer-based architectures \cite{b23,b24,b25}, have been widely applied over the past decade. However, many existing studies rely on older, small scale-datasets, such as CBIS-DDSM \cite{b30} and INbreast \cite{b31}, or only use limited subsets of larger datasets such as OPTIMAM \cite{b18} and EMBED \cite{b19}. This is largely due to restricted access to the complete datasets and the lack of curated lesion level annotations. As a result, the generalization performance of deep learning models on large-scale, multi-site mammography datasets has not been thoroughly tested.

In addition, most prior work primarily focuses on conventional 2D FFDM images, while relatively few studies incorporate or systematically analyze synthetic images as additional inputs \cite{b26,b27,b28,b29}. Synthetic images are derived from digital breast tomosynthesis exams and are increasingly replacing FFDM images in the screening setting to reduce exam time and patient radiation exposure \cite{b41}. Since this is a newer approach, there is a scarcity of large publicly available synthetic image datasets with lesion annotations.

\begin{figure}[!t]
\centerline{\includegraphics[width=\columnwidth]{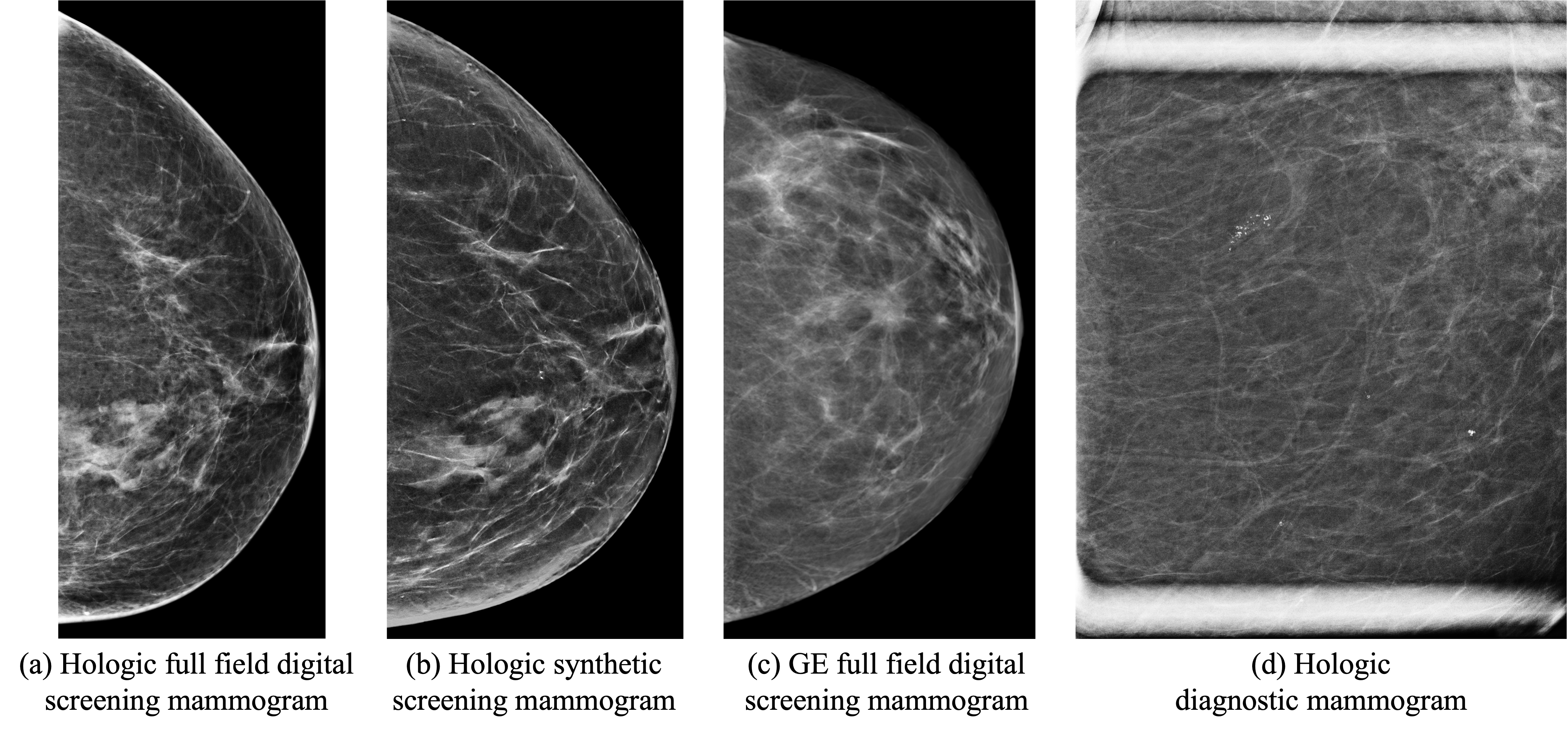}}
\caption{Representative examples of mammograms from different scanner vendors (GE and Hologic) using different techniques (full field digital mammogram and synthetic mammogram; screening and diagnostic), where (a) and (b) are from the same patient, while (c) and (d) are from different patients.}
\label{fig2}
\end{figure}

In this work, to better evaluate model generalization across multi-site datasets, we assembled three large-scale mammography datasets that include both FFDM and synthetic images from multiple vendors (GE and Hologic, Figure \ref{fig2}). We further compared and analyzed both CNN-based and transformer-based deep learning approaches in a unified experimental setting.

\subsection{Style Transfer}
Style transfer has been widely explored as an effective input-level domain adaptation technique. Early work introduced a CNN-based framework that separates and recombines content and style features \cite{b32}. Adaptive instance normalization (AdaIN) aligns content and style features \cite{b16}, while whitening and coloring transform (WCT) enables arbitrary style transfer without retraining for new styles \cite{b33}. With the rapid advancement of generative models, CycleGAN provided an unpaired image-to-image translation framework based on adversarial learning \cite{b17}. More recently, transformer-based structures have been incorporated to better capture global dependencies \cite{b34}, while diffusion models enabled semantic and text-guided style transfer \cite{b35}.

In the medical imaging domain, Synergistic Image and Feature Adaptation (SIFA) combines CycleGAN-based image translation with adversarial feature alignment to reduce domain shift between MR and CT images \cite{b36}. CycleGAN has also been applied to mitigate domain discrepancies arising from different vendors and clinical centers \cite{b37}. In breast imaging, an unsupervised domain adaptation framework with adversarial learning was developed for breast mass detection \cite{b38}. Transformer-based detection models exhibited increased robustness to domain shifts, and MixStyle layers were used to improve generalization in previously unencountered clinical environments \cite{b11}. Finally, contrastive learning has also been applied to encourage domain-invariant feature representations across imaging sources \cite{b39}.

Despite these advances, most existing studies focus on detection or segmentation tasks for breast masses, and only limited work has systematically investigated domain shifts and classification performance for calcifications. Motivated by these approaches, we proposed a multi-stage classification framework that integrates a domain adaptation module with a cascade classification module. Within the domain adaptation stage, we adopted AdaIN and CycleGAN as two representative style transfer-based methods to reduce domain discrepancies across multi-site mammography datasets.

\section{Methods}
\begin{figure}[!t]
\centerline{\includegraphics[width=0.9\columnwidth]{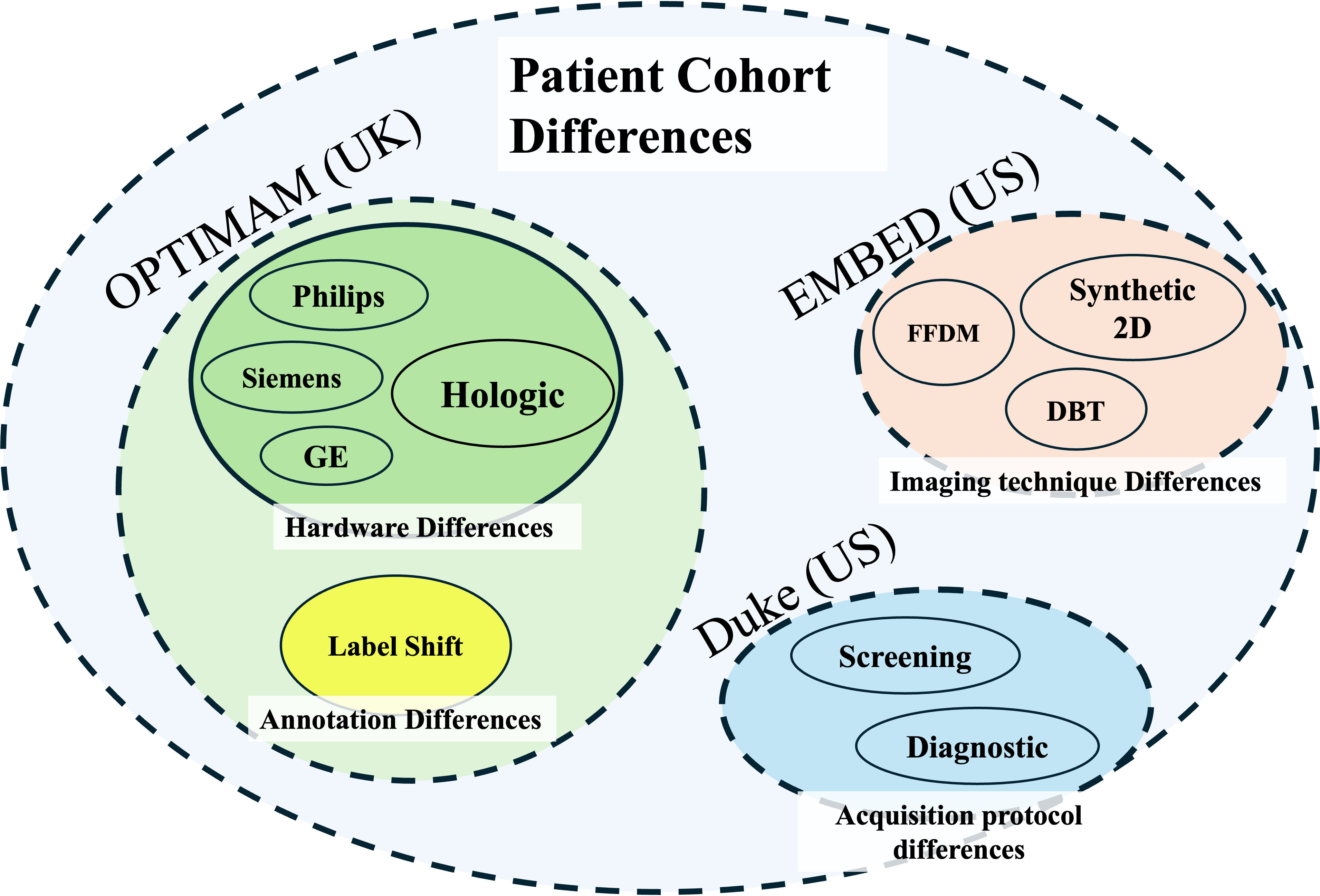}}
\caption{Illustration of domain shifts across multi-site dataset (OPTIMAM, EMBED and Duke Calcification Dataset v1: 2026), including differences across patient cohorts, scanner hardware, annotation practices, imaging techniques and acquisition protocols. UK: United Kingdom, US: United States, FFDM: full field digital mammogram, DBT: digital breast tomosynthesis.}
\label{fig3}
\end{figure}

As summarized in Figure \ref{fig3}, domain shifts in multi-site mammography datasets arise from several sources: (1) hardware differences including variations in mammography systems and scanner manufacturers such as Hologic, Siemens, and GE \cite{b11,b15}; (2) annotation differences reflecting inter-observer variability and annotation protocols; (3) imaging technique differences such as conventional FFDM versus synthetic images reconstructed from digital breast tomosynthesis; (4) protocol differences including variability between screening and diagnostic examinations; and (5) patient cohort differences arising from differences in population demographics and screening intervals between imaging sites, populations, and countries. In this work, we proposed a multi-stage classification framework to improve model generalization focusing on hardware and imaging technique differences.

\begin{figure*}[!t]
\centering
\includegraphics[width=0.8\textwidth]{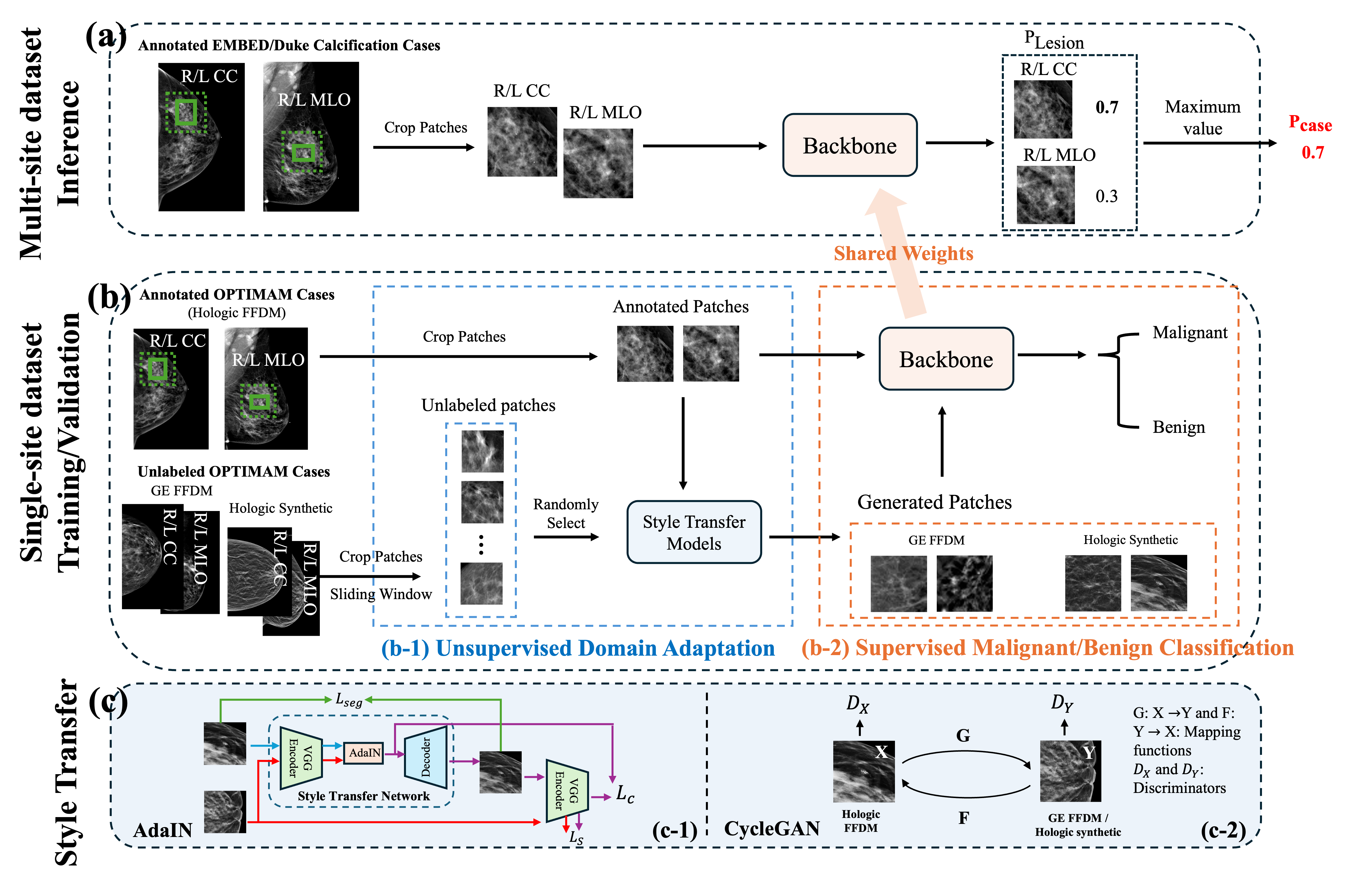}
\caption{Overall framework of the proposed domain adaptation method for calcification classification. The framework included two pipelines: (a) an inference pipeline, which used only the classification backbone network, and (b) a training/validation pipeline, which included an unsupervised domain adaptation module and a supervised malignant/benign classification module. For the single-site classification task, only the supervised backbone module (b-2) was used during training. For multi-site classification, both unsupervised domain adaptation module (b-1) and the supervised classification module (b-2) were used during training. The domain adaptation module applied CycleGAN or AdaIN for style transfer to generate vendor-specific patches while the supervised classification module received both vendor-specific patches and original annotated lesion patches as input. For each module in the proposed framework, only patch-level images were used for training, validation and testing, whereas final classification performance was assessed at the case level. During inference, only the classification module was used.}
\label{fig4}
\end{figure*}

As illustrated in Figure \ref{fig4}, the proposed multi-stage framework consisted of two main components: (b-1) an unsupervised Domain Adaptation Module and (b-2) a supervised malignant/benign classification model. In step (b-1), a style transfer model was trained with both annotated lesion patches from the source dataset and unlabeled vendor-specific patches (e.g. GE FFDM and Hologic synthetic images) to generate lesion patches with corresponding vendor styles. In step (b-2), a classification backbone network was trained to output a malignancy prediction score for each annotated lesion patch. In this study, the OPTIMAM dataset was used as the training dataset for both single-site backbone comparisons and multi-site domain adaptation experiments, while EMBED and the Duke Calcification Dataset v1 were used for inference. 

\subsection{Calcification Classification Within a Single-Site Dataset}
In this work, the OPTIMAM dataset was selected as the single-site dataset source, as it contained the largest number of calcification cases in our study and was publicly available, which will make the results more easily reproducible to other investigators. For single-site classification, only the original annotated lesions from OPTIMAM were used for both training and validation. Each lesion was cropped into patches of size 512x512 pixels. If a lesion exceeded this size in either width or height, a sliding window strategy was applied to extract multiple overlapping patches. No generated images or patches were included at this stage.

Both CNN-based and transformer-based network models were evaluated as classification backbones. Specifically, VGG16, MobileNet\_v2, DenseNet121, ResNet18, ResNet 50, ResNet101 and EfficientNet\_v2\_S were selected as representative CNN-based architectures, converging a wide range of network structures and model complexities that demonstrated strong performance across various computer vision tasks. In addition, Swin Transformer V2 was chosen as the transformer-based backbone. Swin Transformer V2 is a hierarchical vision transformer with multiple stages and shifted window self-attention. It models long-range dependencies while keeping computation scalable for high-resolution images.

\subsection{Domain Adaptation Methods Across Vendors}
In the OPTIMAM dataset, the majority of annotated lesions come from screening FFDM images acquired on Hologic systems. However, in clinical practice, mammograms may be acquired from multiple vendors (e.g., GE, Siemens, Philips), with different image techniques (FFDM and synthetic images) and acquisition protocols (screening and diagnostic with magnification views). To address this mismatch, two style transfer-based domain adaptation methods, AdaIN and CycleGAN, were employed to synthesize vendor-specific image characteristics (GE FFDM and Hologic synthetic images) from unlabeled mammograms in an unsupervised manner.

\textbf{AdaIN} is a feature-level style transfer method that aligns the channel-wise mean and variance between content and style features. It performs one-shot statistical alignment in feature space and reconstructs the stylized output back into pixel space. To better balance preservation of calcification lesion content and incorporation of vendor-specific appearance, an additional segmentation branch was introduced between the original content patch and the AdaIN output. The segmentation branch was based on a U-Net architecture and trained using Dice loss. For each patch, the calcification mask label was generated automatically by a previously trained U-Net model with the same architecture from our earlier work \cite{b42}. The overall loss function was defined as:
\begin{equation}L = L_c + \lambda L_{sty} + \alpha L_{seg} \label{eq}\end{equation}
Where $L_c$ denotes the content loss, $L_{sty}$ is the style loss and $L_{seg}$ is the auxiliary segmentation loss.

\textbf{CycleGAN} is an unpaired image-to-image translation framework consisting of two generators and two discriminators. A cycle-consistency loss was employed to ensure that an image translated to the target domain and then back to the source domain preserved its original structure, enabling effective domain translation without paired training data.

\subsection{Downstream Task: Malignant Versus Benign Classification Across Multi-Site Datasets}
\textbf{Training and Validation:} During training and validation, only images from the OPTIMAM dataset were used. In contrast to the single-site classification stage, the training data now included both the original annotated lesion patches and the generated patches produced by the domain adaptation module. These combined data were used to train the final supervised malignant versus benign classification model.

As shown in Table \ref{tab1:dataset_description}, the proportion of available benign cases in the OPTIMAM dataset is relatively low compared to the other datasets and real-world clinical practice.	To alleviate this imbalance, the ratio of malignant to benign patches was set to 1:2 in each training batch. The entire model remained trainable, and the best model checkpoint was selected based on batch-level classification AUC.

\begin{table*}[!t]
\centering
\caption{Overview of the Datasets Used in This Study}
\label{tab1:dataset_description}
\renewcommand{\arraystretch}{1.15}
\setlength{\tabcolsep}{4pt}

\resizebox{\textwidth}{!}{
\begin{tabular}{c|c|c|c|c|c|c|c|c}
\hline\hline
\multicolumn{9}{c}{\textbf{TABLE 1A: Dataset Description of Multi-site Malignant Versus Benign Classification}} \\
\hline
Dataset & Origin & 
\begin{tabular}[c]{@{}c@{}}Public /\\ Private\end{tabular} &
\begin{tabular}[c]{@{}c@{}}Vendors\\ Included\end{tabular} &
Image Techniques &
\begin{tabular}[c]{@{}c@{}}\# of\\ Total\\ Cases\end{tabular} &
\begin{tabular}[c]{@{}c@{}}\# of\\ Malignant\\ Cases\end{tabular} &
\begin{tabular}[c]{@{}c@{}}\# of\\ Benign\\ Cases\end{tabular} &
\begin{tabular}[c]{@{}c@{}}Train /\\ Validation /\\ Test Split for\\ Classification\end{tabular} \\
\hline

OPTIMAM & UK &
\begin{tabular}[c]{@{}c@{}}Public,\\ permission\\ required\end{tabular} &
Hologic &
\begin{tabular}[c]{@{}c@{}}100.0\% FFDM\\ (100.0\% screening)\end{tabular} &
2994 &
\begin{tabular}[c]{@{}c@{}}2187\\ (73.0\%)\end{tabular} &
\begin{tabular}[c]{@{}c@{}}807\\ (27.0\%)\end{tabular} &
\begin{tabular}[c]{@{}c@{}}Train \&\\ validation\\ (5-fold\\ cross-validation)\end{tabular} \\
\hline

EMBED & US &
\begin{tabular}[c]{@{}c@{}}Only 20\%\\ (the public\\ portion) is\\ available\end{tabular} &
Hologic &
\begin{tabular}[c]{@{}c@{}}100.0\% FFDM\\ (100.0\% screening)\end{tabular} &
125 &
\begin{tabular}[c]{@{}c@{}}35\\ (28.0\%)\end{tabular} &
\begin{tabular}[c]{@{}c@{}}90\\ (72.0\%)\end{tabular} &
\begin{tabular}[c]{@{}c@{}}Independent\\ test\\ (5 models)\end{tabular} \\
\hline

\begin{tabular}[c]{@{}c@{}}Duke\\ Calcification\\ Dataset v1\end{tabular} &
US &
\begin{tabular}[c]{@{}c@{}}Private,\\ in-house\end{tabular} &
\begin{tabular}[c]{@{}c@{}}Hologic,\\ GE\end{tabular} &
\begin{tabular}[c]{@{}c@{}}58.6\% FFDM, 41.4\%\\ synthetic images; 62.5\%\\ screening, 37.5\% diagnostic\\ (26.9\% magnification view)\end{tabular} &
788 &
\begin{tabular}[c]{@{}c@{}}247\\ (31.3\%)\end{tabular} &
\begin{tabular}[c]{@{}c@{}}541\\ (68.7\%)\end{tabular} &
\begin{tabular}[c]{@{}c@{}}Independent\\ test\\ (5 models)\end{tabular} \\
\hline

\multicolumn{9}{c}{\textbf{TABLE 1B: Dataset Description for Domain Adaptation}} \\
\hline
Dataset &
\begin{tabular}[c]{@{}c@{}}Vendors\\ Included\end{tabular} &
\begin{tabular}[c]{@{}c@{}}Image\\ Techniques\end{tabular} &
\# of Cases &
\# of Images &
\# of patches &
\begin{tabular}[c]{@{}c@{}}Lesion\\ Annotation\\ Availability\end{tabular} &
\multicolumn{2}{c}{Train / Validation / Test Split} \\
\hline

OPTIMAM & Hologic & FFDM & 2994 & -- & 6159 & Yes & \multicolumn{2}{c}{Train \& Validation} \\
\hline
OPTIMAM & GE & FFDM & 857 & 2462 & 32323 & No & \multicolumn{2}{c}{Train} \\
\hline
OPTIMAM & Hologic & Synthetic & 275 & 614 & 6014 & No & \multicolumn{2}{c}{Train} \\
\hline\hline

\end{tabular}
}
\end{table*}

\textbf{Inference:} During final inference, only the trained supervised classification model was applied to lesions from independent multi-site datasets. No generated images or patches were used at this stage. For each case, the maximum malignancy score across all extracted lesion patches was taken as the final case-level prediction,
\begin{equation}P_{case} = \max(P_{patch_1}, P_{patch_2}, \ldots, P_{patch_n}) \label{eq}\end{equation}

\subsection{Datasets}
This retrospective study was compliant with the Health Insurance Portability and Accountability Act and received an exemption from the need for informed consent by an institutional review board. As summarized in Table \ref{tab1:dataset_description}A, we used three large-scale curated mammography datasets in this study, including two public datasets (OPTIMAM and EMBED) and one private in-house dataset (Duke Calcification Dataset v1: 2026). Across all datasets, only biopsy-proven calcification cases were included, and each case was labeled as either malignant or benign. Final labeling was assigned based on the initial core needle biopsy and subsequent surgical excision, when performed.  If malignancy was found on either pathology sample, the case was considered malignant, otherwise it was considered benign. DCIS was categorized as malignant.

\textbf{OPTIMAM} is a large, public mammography dataset from the United Kingdom, curated from multiple clinical sites and widely used for breast cancer detection and classification research since 2014 \cite{b18}. In this study, OPTIMAM was used as the sole dataset for model training and validation. Among the annotated mammograms in OPTIMAM, 88.7\% are acquired using Hologic FFDM systems. A total of 2,994 annotated calcification cases were included, comprising 2,187 malignant and 807 benign cases. To support unsupervised domain adaptation, we additionally included 431 GE FFDM images and 614 Hologic synthetic images, yielding over 30,000 unlabeled patches in total. For these unlabeled images, Otsu’s method was first applied to segment the breast region from the original mammograms, followed by a sliding-window approach with 128-pixel overlap to crop the entire breast area into 512 × 512-pixel patches. All images were selected from pre-biopsy exams containing mass lesions rather than normal or existing calcification lesions to introduce more diverse and complex tissue patterns. Compared with normal patches, mass-case patches provide more realistic and challenging tissue background variations that better reflect clinical practice. This helped better capture cross-domain variability and may improve the learning of domain-invariant features and model robustness to domain shifts. Details of annotation availability and the training/validation split are provided in Table \ref{tab1:dataset_description}.

\textbf{EMBED} is a large mammography dataset from Emory University, of which 20\% of the FFDM and synthetic exams are publicly available for research \cite{b19}. In this study, we included 125 biopsy-proven exams with annotated calcifications from the EMBED dataset. Similar to OPTIMAM, most annotated mammograms in EMBED are Hologic FFDM images. EMBED was used exclusively as an independent test dataset for multi-site evaluation and was not involved in any training or validation procedures.

\textbf{Duke Calcification Dataset v1} is a private U.S.-based mammography dataset collected at Duke University Medical Center between 2008 and 2023. It contains 788 biopsy-proven calcification cases, with 37.1\% of images acquired from GE systems and 62.9\% from Hologic systems. As summarized in Table \ref{tab1:dataset_description}A, the dataset includes both screening (62.5\%) and diagnostic (37.5\%) examinations. Among the diagnostic exams, 26.9\% of images are magnification views. Overall, 58.6\% of the dataset consists of FFDM images and 41.4\% consists of synthetic images, indicating a relatively higher proportion of magnification and synthetic images compared with the public datasets. Lesion-level annotations were manually performed by a fellowship trained breast radiologist (11 years of experience) with radiology reports used as reference. This dataset was used solely as an independent test set to evaluate model generalization across multi-site datasets.

\subsection{Implementation Details}
In both the single-site and multi-site classification stages, each annotated lesion was cropped into patches of size 512 × 512 pixels during training, validation, and testing. For the OPTIMAM dataset, a five-fold cross-validation scheme was used during training, and the final receiver operating characteristic (ROC) and area under the curve (AUC) was reported as the average across all five validation folds.

All network structures in this study were implemented using Pytorch 2.4 and trained on four NVIDIA 2080 Ti GPUs. Classification backbone networks were initialized with ImageNet-pretrained weights using Adam optimizer with a learning rate of 0.00005. For the domain adaptation models, we followed the original configurations described in their respective publications: VGG19 was used as the encoder for AdaIN; CycleGAN employed a ResNet with 9 blocks of 20 convolution layers.

\section{EXPERIMENTS AND RESULTS}

\subsection{Comparison Within Single-Site Dataset}
\begin{figure}[!t]
\centerline{\includegraphics[width=\columnwidth]{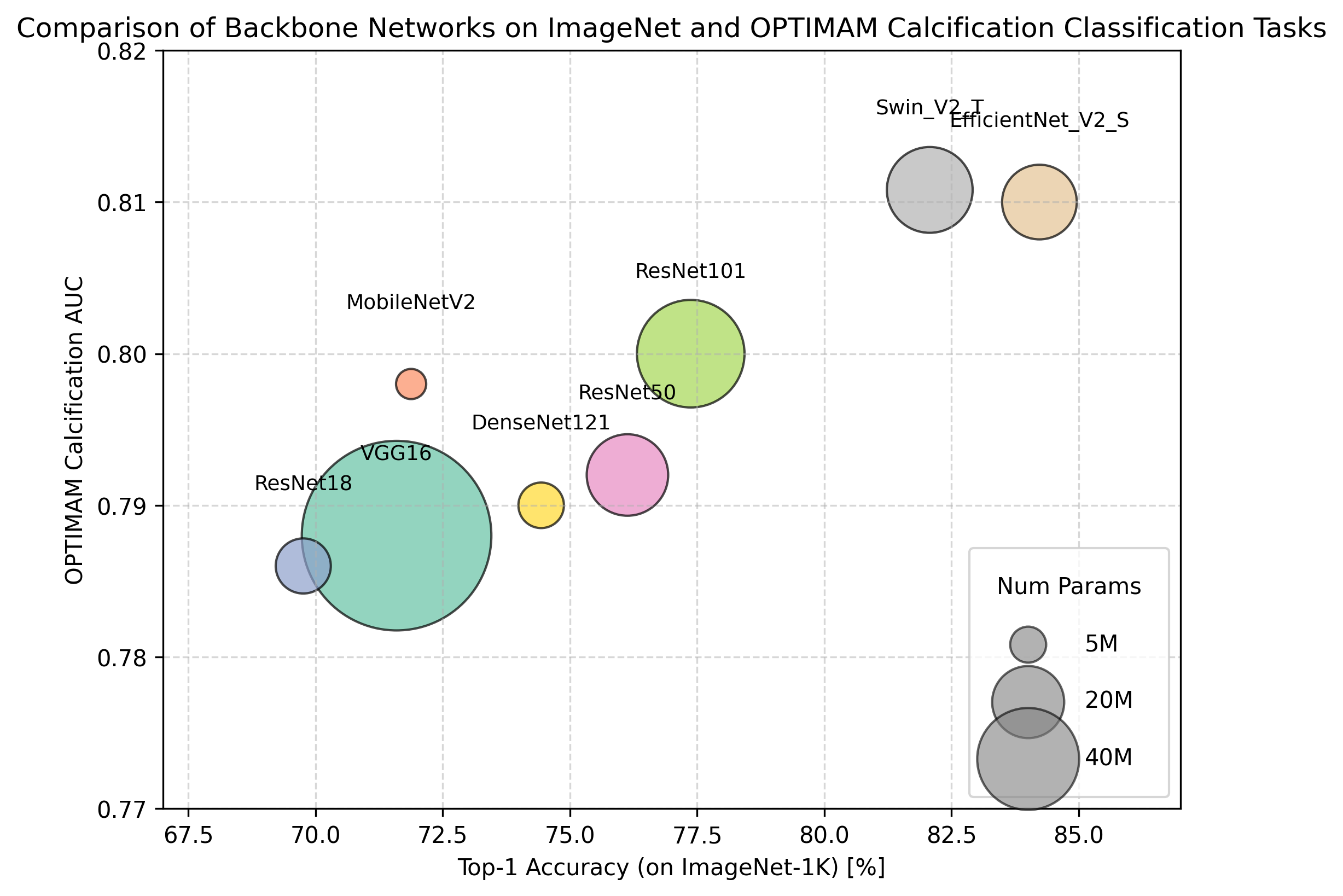}}
\caption{Performance of seven CNN-based backbone networks and a Swin Transformer-based backbone. Although the pre-trained backbones differed greatly on ImageNet-1k, fine-tuned calcification classification performance was similar (AUC 0.78-0.81). Swin Transformer V2 was chosen as the baseline because it achieved the highest AUC (0.81) with moderate model complexity.}
\label{fig5}
\end{figure}

We first evaluated seven CNN-based backbone networks and one transformer-based backbone to identify the most suitable model for malignant versus benign calcification classification on the single-site OPTIMAM dataset. As shown in Figure \ref{fig5}, although these pre-trained backbone networks exhibited substantial differences in Top-1 accuracy on the ImageNet-1k benchmark \cite{b40} (ranging from approximately 70\% to 85\% in accuracy), their fine-tuned performance on calcification classification was comparable, with AUC ranging from 0.78 to 0.81.

Based on these results, we selected Swin Transformer V2 as the baseline backbone for downstream experiments. Swin Transformer V2 achieved the highest AUC of 0.81 in binary calcification classification (benign vs. malignant), while maintaining moderate model complexity. In addition, its hierarchical architecture and scalability made it well suited to leverage larger calcification datasets in subsequent multi-site evaluations.

\begin{figure}[!t]
\centerline{\includegraphics[width=\columnwidth]{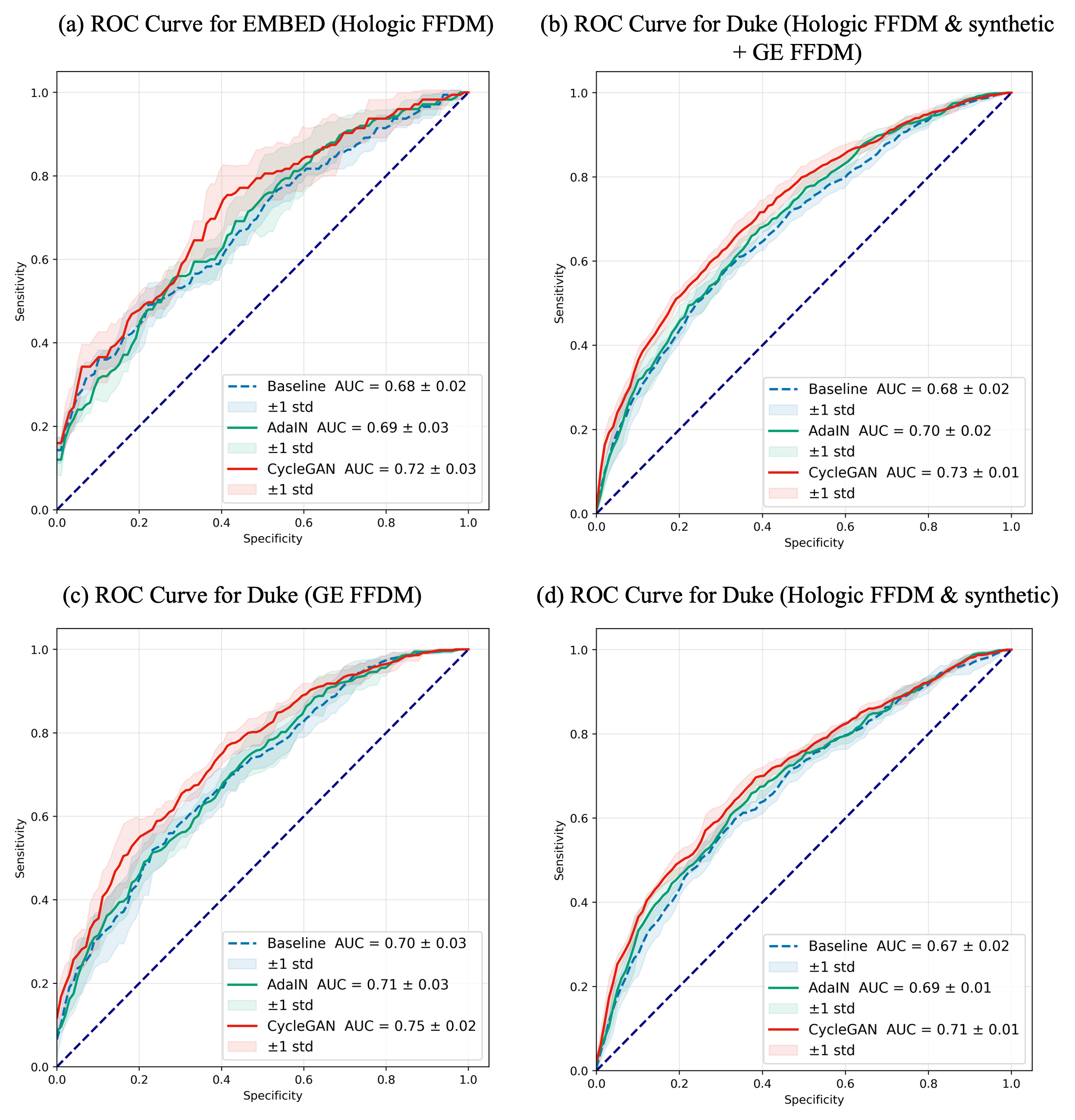}}
\caption{Comparison of OPTIMAM-trained baseline without domain adaptation versus domain adaptation-based framework in external validations with EMBED and Duke datasets (Duke Calcification Dataset v1). The full framework with CycleGAN achieves consistently higher sensitivity at corresponding specificity operating points on both external datasets.}
\label{fig6}
\end{figure}

\begin{figure*}[!t]
\centering
\includegraphics[width=\textwidth]{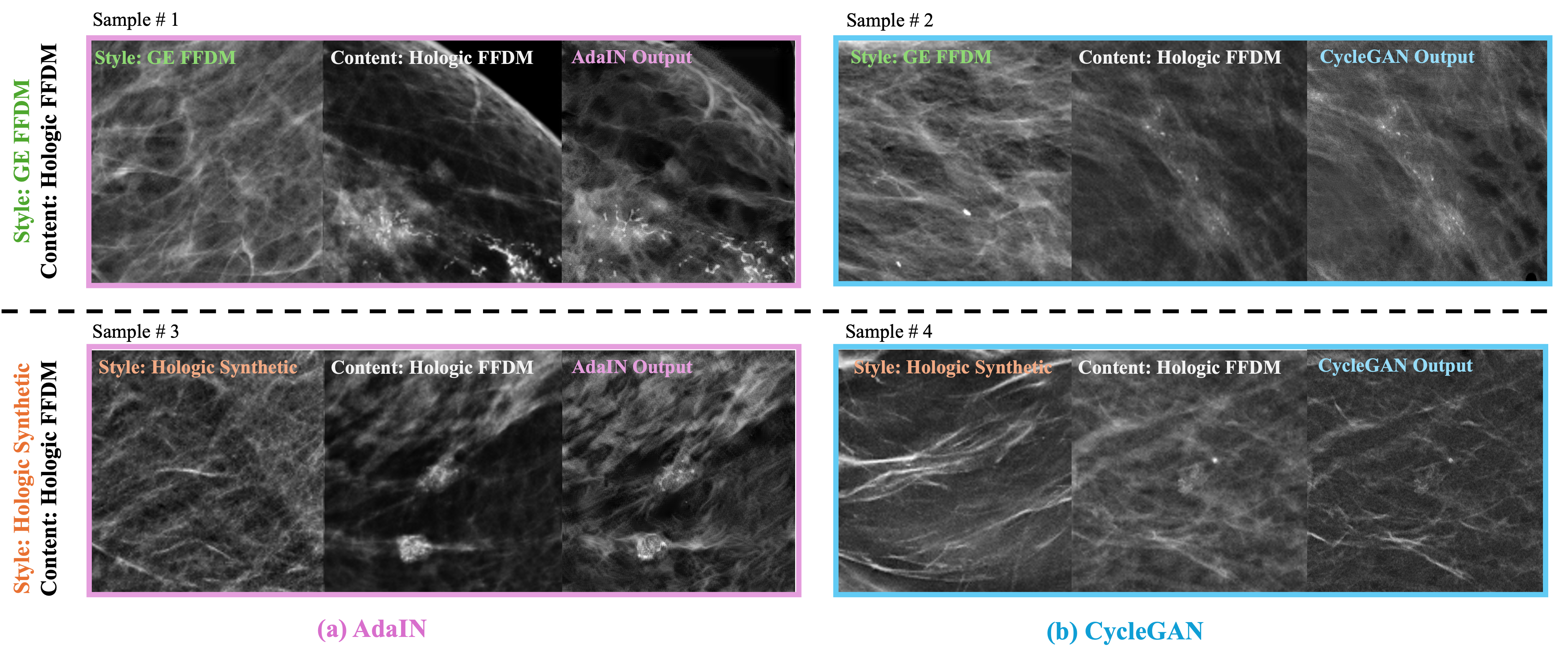}
\caption{Visual comparison of style transfer results. GE FFDM and Hologic synthetic patches were used as style patches in the first and second rows, respectively, while Hologic FFDM patches are used as content patches in both rows. The first and second columns show results from AdaIN and CycleGAN, respectively.}
\label{fig7}
\end{figure*}

\subsection{Visual Comparison of Domain Adaptation Model Outputs}
As shown in Figure \ref{fig7}, given randomly selected style patches (GE FFDM or Hologic synthetic) and annotated calcification patches, both AdaIN and CycleGAN generate outputs with vendor-specific and technique-specific appearance characteristics. For example, the brightness and sharpness of the original patches are modified after style transfer. However, AdaIN tends to introduce artificial connections between individual calcifications and produces relatively blurred regions surrounding the lesions. In comparison, CycleGAN preserves clearer calcification boundaries without introducing false lesion-like structures. 

\subsection{Comparison Across Multi-Site Datasets}
Next, we evaluated the full multi-site calcification classification framework. As shown in Table \ref{tab2:comparison_results}, incorporating domain adaptation methods yielded clear improvements in downstream calcification classification performance, as measured by AUC. In particular, the use of both the original annotated lesion patches and CycleGAN-generated patches increased the AUC from 0.68 to 0.72 on the EMBED dataset and from 0.68 to 0.73 on the Duke Calcification Dataset v1, corresponding to relative gains exceeding 5\% in both cases. Consistent performance improvements were also observed across vendors within the Duke Calcification Dataset v1. For images acquired from GE systems, the AUC increased from 0.70 to 0.75, while for Hologic images, it improved from 0.67 to 0.71.

In addition, there was a consistently higher sensitivity at corresponding specificity operating points, for both the EMBED dataset and the Duke Calcification Dataset v1 (Figure \ref{fig6}).

\begin{table*}[!t]
\centering
\caption{Comparison Results on the Validation and Independent Test Datasets, Where Duke Refers to Duke Calcification Dataset v1. All Values are Reported as Mean $\pm$ Standard Deviation, and Bold Values Indicate the Highest AUC Achieved by the Proposed Method.}
\label{tab2:comparison_results}
\renewcommand{\arraystretch}{1.15}
\setlength{\tabcolsep}{6pt}

\resizebox{\textwidth}{!}{
\begin{tabular}{c|c|c|c|c|c|c}
\hline\hline
\begin{tabular}[c]{@{}c@{}}Unsupervised\\ Domain\\ Adaptation\end{tabular} &
\begin{tabular}[c]{@{}c@{}}Domain\\ Adaptation\\ Methods\end{tabular} &
\begin{tabular}[c]{@{}c@{}}Train /\\ Validation\\ Dataset (5-fold\\ Cross-validation)\end{tabular} &
\begin{tabular}[c]{@{}c@{}}Independent\\ Test\\ Dataset\end{tabular} &
\multicolumn{3}{c}{AUC} \\
\cline{5-7}
 &  &  &  &
\begin{tabular}[c]{@{}c@{}}Hologic Only\\ (FFDM +\\ Synthetic)\end{tabular} &
\begin{tabular}[c]{@{}c@{}}GE Only\\ (FFDM)\end{tabular} &
All \\
\hline

 &  & OPTIMAM &  & $0.81 \pm 0.01$ &  &  \\
\hline
 &  & OPTIMAM & EMBED & $0.68 \pm 0.02$ &  &  \\
\hline
$\checkmark$ & AdaIN & OPTIMAM & EMBED & $0.69 \pm 0.03$ &  &  \\
\hline
$\textbf{\checkmark}$ & \textbf{CycleGAN} & \textbf{OPTIMAM} & \textbf{EMBED} & $\mathbf{0.72 \pm 0.03}$ &  &  \\
\hline

 &  & OPTIMAM & Duke & $0.67 \pm 0.02$ & $0.70 \pm 0.03$ & $0.68 \pm 0.02$ \\
\hline
$\checkmark$ & AdaIN & OPTIMAM & Duke & $0.69 \pm 0.01$ & $0.71 \pm 0.03$ & $0.70 \pm 0.02$ \\
\hline
$\textbf{\checkmark}$ & \textbf{CycleGAN} & \textbf{OPTIMAM} & \textbf{Duke} & $\mathbf{0.71 \pm 0.01}$ & $\mathbf{0.75 \pm 0.02}$ & $\mathbf{0.73 \pm 0.01}$ \\
\hline\hline
\end{tabular}
}
\end{table*}

\section{Discussion}
This study focused on the problem of malignant versus benign calcification classification across multi-site mammography datasets. Our results show that although current deep learning models can achieve reasonable performance within a single dataset, their generalization to external datasets remains limited due to domain shifts. This is an important challenge in clinical practice, where medical imaging data are collected across different institutions, vendors, and imaging workflows.

Our single-site experiments on OPTIMAM showed that different backbone networks achieved similar performance, despite large differences in their ImageNet benchmark accuracy. Previous work has shown that stronger performance on natural image classification does not necessarily translate into better performance in medical imaging \cite{b43}, and our study supports this finding for calcification classification in mammography. The multi-site evaluation further demonstrated the impact of domain shifts on this task. When the baseline model trained on OPTIMAM was directly applied to the other two datasets, EMBED and the Duke Calcification Dataset v1, the performance dropped noticeably. After incorporating the proposed domain adaptation module, which generates vendor-specific patches without requiring additional annotations, classification performance improved consistently across external datasets. These findings indicate that unseen domain differences are non-trivial for calcification classification and support our hypothesis that style transfer-based augmentation can reduce domain discrepancies in unseen domains and improve model robustness.

In addition to domain shifts introduced by differences in vendor and imaging technique, case availability also contributed to class imbalance. In this study, OPTIMAM was selected as the sole training dataset because it contains the largest number of calcification cases. However, its class distribution is overweighted with malignant cases. By contrast, the other two datasets contain more benign cases than malignant cases, a distribution that better matches real-world clinical settings. To alleviate this label shift caused by annotation differences, we explored several resampling strategies, including random oversampling and undersampling, but neither improved performance over the baseline. In the undersampling experiments, we randomly reduce the number of malignant cases to either match or halve the number of benign cases, but external validation performance decreased due to the substantial reduction in training data. We found that using more negative samples than positive samples within each training batch, together with incorporating additional generated patches from the style transfer models, was a more effective strategy and led to the best classification performance.

To the best of our knowledge, this is the first study to use both the publicly available OPTIMAM and EMBED datasets specifically for the task of malignant versus benign calcification classification, while also leveraging the largest calcification subset available in OPTIMAM. Previous calcification classification studies using OPTIMAM generally relied on relatively small case samples \cite{b44,b45}. Other classification studies have primarily focused on masses or on mixed masses with calcifications, where mass lesion findings dominated the easier classification task \cite{b46,b47,b48,b49}. Moreover, many studies involving these datasets have investigated different tasks, such as detection \cite{b53,b54} and risk prediction \cite{b50,b51,b52}. These differences make direct comparison with our results challenging.

This work has several limitations. First, AdaIN and CycleGAN were selected as two representative style transfer strategies that are well-established methods. The AdaIN framework is simple and efficient, while CycleGAN provides a more flexible adversarial approach for unpaired image-to-image translation that is more flexible. However, both are relatively older methods, and more advanced approaches, such as diffusion-based methods, may provide better image quality and more stable domain translation. Second, although OPTIMAM included around 3000 cases, it was still not large enough to fully avoid bias from a single training and validation split. Therefore, we used five-fold cross-validation, resulting in five trained models, and reported the average performance on the external datasets. While this provides a more robust evaluation, it is less practical for clinical deployment, where a single model would be easier to interpret and use.

\section{Conclusion}
In this work, we presented a multi-stage framework for calcification classification which addresses domain shifts across multi-site mammography datasets using different vendors and imaging techniques. By combining an unsupervised domain adaptation module with a supervised classifier, the proposed method achieved consistent improvements across external datasets, showing that unsupervised domain adaptation can effectively reduce cross-site performance gap. Our future work will extend this framework to additional CAD tasks and other lesion types.

\end{document}